\documentclass[conference]{IEEEtran}
\IEEEoverridecommandlockouts
\usepackage{amsmath,amssymb,amsfonts}
\usepackage{algorithmic}
\usepackage{graphicx}
\usepackage{textcomp}
\usepackage{xcolor}
\usepackage{subfig}

\usepackage[bibstyle=ieee,citestyle=numeric-comp,sortcites,mincitenames=1,maxcitenames=2,natbib=true,backend=biber]{biblatex}
\def\BibTeX{{\rm B\kern-.05em{\sc i\kern-.025em b}\kern-.08em
    T\kern-.1667em\lower.7ex\hbox{E}\kern-.125emX}}

\title{Demonstration-Based Robotic Feeding Using Dynamic Movement Primitives}
\author{Kenzhi Iskandar Wong$^1$, Kitti Thamrongaphichartkul$^1$, Sri Harsha Turlapati$^2$, Yanshuo Lu$^1$, Kanichi Doi$^3$, \\ Boyu Ma$^1$, Bin Zhou$^1$, Lin Yang$^1$, Ananda Sidarta$^4$, Lek Syn Lim$^4$, Arun-Kumar Kaliya-Perumal$^4$, \\ Jianfei Yang$^1$, Domenico Campolo${^{1,4,\ast}}$
\thanks{$^{1}$Robotics Research Centre, School of Mechanical
and Aerospace Engineering, Nanyang Technological University, Singapore.}%
\thanks{$^{2}$School of Science and Technology, Singapore University of Social Sciences, Singapore.}
\thanks{$^{3}$Department of Mechano-Informatics, University of Tokyo, Japan.}
\thanks{$^{4}$R3 Institute Singapore, Nanyang Technological University, Singapore.}
\thanks{Open access funding support was provided by the Rehabilitation Research Institute of Singapore (Grant ID: 021099-00001), a tripartite collaboration between Nanyang Technological University (NTU), the Agency for Science, Technology and Research (A*STAR), and NHG Health.}
\thanks{$^{\ast}$Corresponding author: {\tt\small d.campolo@ntu.edu.sg}}
}
\date{} % This removes the date

\begin{document}
\maketitle

%%%%%%%%%%%%%%%%%%%%%%%%%%%%%%%%%%%%%%%%%%%%%%%%%%%%%%%%%%%%%%%%%%%%%%%%%%%%%%%%%%%%%%
\begin{abstract}
This paper presents a demonstration-based robotic feeding framework that
integrates human demonstration, Dynamic Movement Primitive-based motion
planning, vision-based scene perception, and robot control. A human
operator first demonstrates the scooping and feeding motions while the
position and Euler-angle orientation of the spoon are recorded using a
motion-capture system. The recorded trajectory is segmented into
scooping and feeding motions, which are encoded using separate Dynamic
Movement Primitives. Before task execution, an RGB-D camera estimates
the participant's mouth position, while two AprilTags are used for
robot-frame registration and tray localization. The detected tray pose
is used to position the scooping motion, whereas the detected mouth
position defines the positional goal of the feeding motion. The
approach, scooping, and feeding trajectories are then executed
sequentially by a Kinova Gen3 manipulator through a ROS~2-based
framework. A real-world experiment demonstrates the feasibility of the
integrated pipeline and shows that the learned spoon motions can be
reproduced within the detected experimental scene.
\end{abstract}


%%%%%%%%%%%%%%%%%%%%%%%%%%%%%%%%%%%%%%%%%%%%%%%%%%%%%%%%%%%%%%%%%%%%%%%%%%%%%%%%%%%%%%
\section{Introduction}
\label{sec:introduction}

Robot-assisted feeding aims to support individuals with upper-limb motor
impairments while reducing reliance on continuous caregiver assistance
\cite{bhattacharjee2019community,park2020active}.
A robotic feeding system must coordinate several components within a
complete task sequence, including food acquisition, utensil transport,
mouth localization, motion generation, and robot control.

% Previous studies have addressed different aspects of robot-assisted
% feeding. End-to-end systems have integrated food acquisition, bite
% transfer, user interfaces, and safety mechanisms to support autonomous
% or semi-autonomous feeding
% \cite{park2020active,gordon2024adaptable}. Other works have investigated
% bite-transfer trajectories that balance efficiency and user comfort
% \cite{belkhale2022balancing}, visual and haptic sensing during
% inside-mouth food transfer \cite{shaikewitz2023mouth}, and robust mouth
% perception combined with physical-interaction-aware control
% \cite{jenamani2024feel}. More recent systems have considered flexible
% mealtime assistance across different activities, environments, and user
% preferences \cite{jenamani2025feast}. Related studies in broader robotic
% assistance and manipulation have explored human-motion retargeting for
% robot control \cite{salem2026pose}, compliant human--robot interaction
% for rehabilitation \cite{merkuryev2025robust}, and compliant actuation
% for robotic manipulation \cite{hoang2026dynamic}. Together, these works
% highlight the importance of perception, motion generation, and compliant
% control in robotic systems that physically interact with users and their
% environment.

Previous studies have addressed different components of robot-assisted
feeding, including end-to-end feeding systems
\cite{park2020active}, bite-transfer trajectories
considering efficiency and user comfort \cite{belkhale2022balancing},
visual and haptic sensing for inside-mouth food transfer
\cite{shaikewitz2023mouth}, and robust mouth perception with
physical-interaction-aware control \cite{jenamani2024feel}. Flexible
mealtime assistance across different activities and user preferences has
also been investigated \cite{jenamani2025feast}. Related robotic systems
have explored human-motion retargeting, compliant rehabilitation
control, and compliant manipulation
\cite{salem2026pose,merkuryev2025robust,hoang2026dynamic}.

An important component of the feeding task is the generation of the
spoon motion. Scooping and feeding require task-specific trajectories in
which both the position and orientation of the spoon evolve throughout
the motion. Learning from demonstration provides a practical means of transferring human-demonstrated motions to robotic systems \cite{pastor2009learning}, as well as task-relevant skills and interaction strategies \cite{golani2025haptic}. Among the available representations, Dynamic
Movement Primitives (DMPs) are well suited for encoding demonstrated
motions as stable nonlinear dynamical systems shaped by learned forcing
terms \cite{ijspeert2013dynamical,saveriano2023dynamic}. DMP-based approaches have also been applied and extended to various manipulation tasks, including contact-rich and interaction-aware scenarios \cite{yang2025planning1,yang2025planning2,nguyen2025quasi}.

Despite these advances, transferring demonstrated spoon motions into a
complete feeding pipeline that adapts them to the current task scene
remains a challenge. Direct replay of a demonstrated trajectory ties the
robot motion to the conditions under which the demonstration was
recorded, whereas a practical feeding system should adapt the learned
scooping and feeding motions to the current bowl and mouth locations.
This paper addresses this problem through an integrated framework that
combines human demonstration, DMP-based motion generation, RGB-D-based
mouth localization, AprilTag-based tray localization, and robotic
execution. The learned scooping motion is retargeted according to the
detected tray location, while the detected mouth position defines the
feeding goal. The scope of the present work covers offline motion
learning and task execution using scene information acquired before the
robot begins moving. Continuous mouth tracking and online adaptation,
food detection, and more complex food-manipulation tasks requiring
cutting or multiple utensils are outside the current scope.

The remainder of this paper is organized as follows.
Section~\ref{sec:preliminaries} introduces the DMP formulation and
learning procedure. Section~\ref{sec:framework} describes the
demonstration-based feeding framework, mouth perception, task
execution sequence, and robot control. Section~\ref{sec:experiment} presents the
experimental setup and results. Section~\ref{sec:discussion} discusses the main findings and limitations, and Section~\ref{sec:conclusion} concludes the paper.

%%%%%%%%%%%%%%%%%%%%%%%%%%%%%%%%%%%%%%%%%%%%%%%%%%%%%%%%%%%%%%%%%%%%%%%%%%%%%%%%%%%%%%
\section{Preliminaries}
\label{sec:preliminaries}

\subsection{Dynamic Movement Primitives}
\label{subsec:dmp}

Dynamic movement primitives (DMPs) represent trajectories as stable
nonlinear dynamical systems consisting of a point-attractor system whose
motion is shaped by a learnable nonlinear forcing term
\cite{ijspeert2013dynamical}. The forcing term encodes the shape of a
demonstrated trajectory, while the initial configuration, goal
configuration, and execution duration can be modified during reproduction.

In this work, the end-effector configuration is represented in a
six-dimensional Euclidean space as
\begin{equation}
    \mathbf{y}(t)
    =
    \begin{bmatrix}
        \mathbf{p}^{\top}(t) &
        \boldsymbol{\eta}^{\top}(t)
    \end{bmatrix}^{\top}
    \in\mathbb{R}^{6},
    \label{eq:dmp_configuration}
\end{equation}
where $\mathbf{p}(t)\in\mathbb{R}^{3}$ is the Cartesian position and
\begin{equation}
    \boldsymbol{\eta}(t)
    =
    \begin{bmatrix}
        \phi(t) &
        \theta(t) &
        \psi(t)
    \end{bmatrix}^{\top}
    \in\mathbb{R}^{3}
\end{equation}
is the Euler-angle vector containing the roll, pitch, and yaw angles,
respectively.

The DMP transformation system is defined as
\begin{align}
    \tau\dot{\mathbf{z}}
    &=
    \alpha_z
    \left[
        \beta_z
        \left(
            \mathbf{g}-\mathbf{y}
        \right)
        -
        \mathbf{z}
    \right]
    +
    \mathbf{A}\mathbf{f}(s),
    \label{eq:dmp_transformation_z}
    \\
    \tau\dot{\mathbf{y}}
    &=
    \mathbf{z},
    \label{eq:dmp_transformation_y}
\end{align}
where $\mathbf{y},\mathbf{z},\mathbf{g}\in\mathbb{R}^{6}$ denote the
current configuration, scaled generalized velocity, and goal
configuration, respectively. The parameter $\tau>0$ temporally scales
the motion, while $\alpha_z>0$ and $\beta_z>0$ determine the convergence
behavior. The scaling matrix is defined as
\begin{equation}
    \mathbf{A}
    =
    \operatorname{diag}
    \left(
        \mathbf{g}-\mathbf{y}_0
    \right)
    \in\mathbb{R}^{6\times6},
    \label{eq:dmp_spatial_scaling}
\end{equation}
where $\mathbf{y}_0\in\mathbb{R}^{6}$ is the initial configuration.
A common choice is $\beta_z=\alpha_z/4$, resulting in critically damped
nominal dynamics.

The motion progression is governed by the canonical system
\begin{equation}
    \tau\dot{s}
    =
    -\alpha_s s,
    \qquad
    s(0)=1,
    \label{eq:dmp_canonical}
\end{equation}
where $s\in\mathbb{R}$ is the phase variable and $\alpha_s>0$ is a
constant. The phase variable monotonically decreases from one toward
zero as the motion progresses.

The nonlinear forcing term
$\mathbf{f}(s)\in\mathbb{R}^{6}$ shapes the generated trajectory and is
defined as
\begin{equation}
    \mathbf{f}(s)
    =
    \frac{
        \displaystyle\sum_{i=1}^{N}
        \mathbf{w}_i\psi_i(s)
    }{
        \displaystyle\sum_{i=1}^{N}
        \psi_i(s)
    }s,
    \label{eq:dmp_forcing}
\end{equation}
where $N$ is the number of radial basis functions and
$\mathbf{w}_i\in\mathbb{R}^{6}$ is the weight vector associated with
the $i$th basis function. Each basis function is given by
\begin{equation}
    \psi_i(s)
    =
    \exp
    \left[
        -h_i(s-c_i)^2
    \right],
    \label{eq:dmp_basis}
\end{equation}
where $c_i\in\mathbb{R}$ specifies the center of the $i$th Gaussian
basis function along the phase variable $s$, and $h_i>0$ controls its
width. A larger value of $h_i$ produces a narrower basis function,
whereas a smaller value produces a wider basis function.

As $s\rightarrow0$, the forcing term vanishes and the transformation
system reduces to a linear spring--damper system that converges to the
goal configuration $\mathbf{g}$.

%%%%%%%%%%%%%%%%%%%%%%%%%%%%%%%%%%%%%%%%%%%%%%%%%%%
\subsection{DMP Learning}
\label{subsec:dmp_learning}

Following the supervised DMP learning procedure in
\cite{ijspeert2013dynamical}, the forcing term is learned from a
demonstrated trajectory
$\mathbf{y}^{d}(t)\in\mathbb{R}^{6}$, where the superscript $d$ denotes
a quantity obtained from the demonstration. Its initial and goal
configurations are defined as
\begin{equation}
    \mathbf{y}_{0}^{d}
    =
    \mathbf{y}^{d}(0),
    \qquad
    \mathbf{g}^{d}
    =
    \mathbf{y}^{d}(T_d),
    \label{eq:dmp_demonstration_boundaries}
\end{equation}
where $T_d$ is the duration of the demonstrated motion. The demonstrated
velocity $\dot{\mathbf{y}}^{d}(t)$ and acceleration
$\ddot{\mathbf{y}}^{d}(t)$ are obtained from the recorded trajectory.

For a constant $\tau$, substituting the demonstrated trajectory into
\eqref{eq:dmp_transformation_z}--\eqref{eq:dmp_transformation_y} and
rearranging the transformation system gives the demonstration-derived
forcing term
\begin{align}
    \mathbf{f}^{d}(t)
    &=
    \left(\mathbf{A}^{d}\right)^{-1}
    \Big[
        \tau^{2}\ddot{\mathbf{y}}^{d}(t)
        \nonumber\\[-1mm]
    &\qquad
        -
        \alpha_z
        \left(
            \beta_z
            \left[
                \mathbf{g}^{d}-\mathbf{y}^{d}(t)
            \right]
            -
            \tau\dot{\mathbf{y}}^{d}(t)
        \right)
    \Big],
    \label{eq:dmp_demonstrated_forcing}
\end{align}
where
\begin{equation}
    \mathbf{A}^{d}
    =
    \operatorname{diag}
    \left(
        \mathbf{g}^{d}-\mathbf{y}_{0}^{d}
    \right)
    \in\mathbb{R}^{6\times6},
    \label{eq:dmp_demonstration_scaling}
\end{equation}
with every element of $\mathbf{g}^{d}-\mathbf{y}_{0}^{d}$ assumed to be nonzero.

The normalized basis-function vector is defined as
\begin{equation}
    \boldsymbol{\varphi}(s)
    =
    \frac{s}{
        \displaystyle\sum_{i=1}^{N}\psi_i(s)
    }
    \begin{bmatrix}
        \psi_1(s) &
        \cdots &
        \psi_N(s)
    \end{bmatrix}^{\top}
    \in\mathbb{R}^{N}.
    \label{eq:dmp_feature_vector}
\end{equation}
Accordingly, the learned forcing term can be expressed as
\begin{equation}
    \mathbf{f}(s)
    =
    \mathbf{W}\boldsymbol{\varphi}(s),
    \qquad
    \mathbf{W}
    =
    \begin{bmatrix}
        \mathbf{w}_1 &
        \cdots &
        \mathbf{w}_N
    \end{bmatrix}
    \in\mathbb{R}^{6\times N},
    \label{eq:dmp_weight_matrix}
\end{equation}
where $\mathbf{W}$ contains the weights of all basis functions.

The learned weights are obtained by fitting
$\mathbf{f}(s)=\mathbf{W}\boldsymbol{\varphi}(s)$ to the
demonstration-derived forcing term $\mathbf{f}^d(t)$ using regularized least-squares
regression:
\begin{equation}
    \mathbf{W}
    =
    \mathbf{F}^{d}\mathbf{\Phi}^{\top}
    \left(
        \mathbf{\Phi}\mathbf{\Phi}^{\top}
        +
        \lambda\mathbf{I}_{N}
    \right)^{-1},
    \label{eq:dmp_weight_learning}
\end{equation}
where
$\mathbf{F}^{d}
=
[\mathbf{f}^{d}(t_1),\ldots,\mathbf{f}^{d}(t_P)]$
and
$\mathbf{\Phi}
=
[\boldsymbol{\varphi}(s_1),\ldots,\boldsymbol{\varphi}(s_P)]$
collect the demonstration-derived forcing terms and basis-function
vectors over the recorded trajectory, respectively, and $\lambda>0$
is a regularization coefficient.
%%%%%%%%%%%%%%%%%%%%%%%%%%%%%%%%%%%%%%%%%%%%%%%%%%%%%%%%%%%%%%%%%%%%%%%%%%%%%%%%%%%%%%
\section{Framework}
\label{sec:framework}

\subsection{System Overview}
\label{subsec:system_overview}

\begin{figure}
    \centering
    \includegraphics[width=\linewidth]{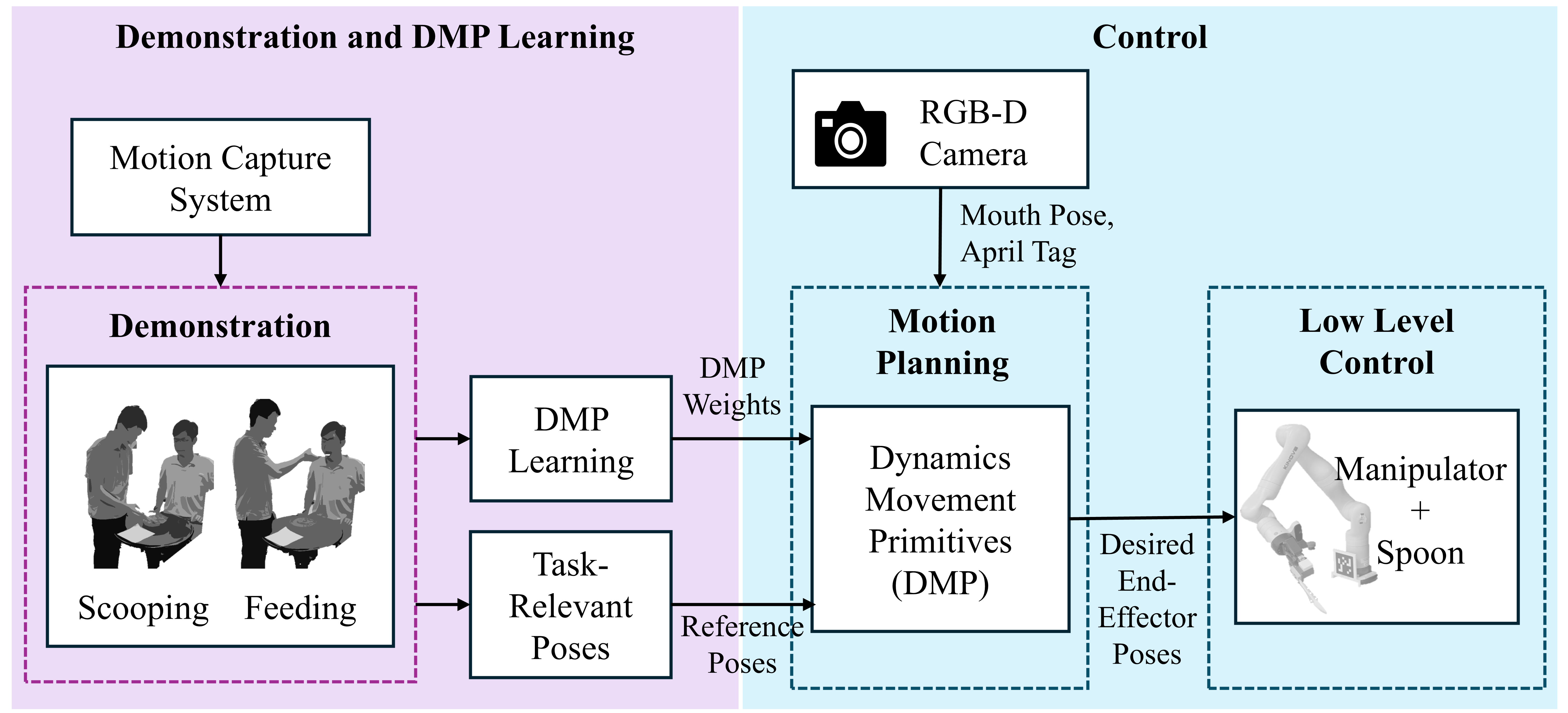}
    \caption{Overview of the demonstration-based robotic feeding framework.
    The scooping and feeding motions are recorded using a motion-capture
    system and encoded as DMPs. Before task execution, the mouth position is
    detected once using an RGB-D camera, while two AprilTags are used for camera-to-robot frame
    registration and tray localization. The learned DMP weights, task-relevant configurations, and
    initial mouth position are used to generate the desired end-effector
    motion for the robotic manipulator.}
    \label{fig:system_overview}
\end{figure}

The framework is implemented using ROS~2 and consists of two main stages: demonstration and DMP
learning, followed by robotic task execution, as illustrated in
Fig.~\ref{fig:system_overview}. In the first stage, a human operator
demonstrates the scooping and feeding motions while the spoon trajectory
is recorded using a motion-capture system. The demonstrations are then segmented into the scooping and feeding
motions, from which the corresponding DMP weights and the initial and
final poses of each motion segment are extracted.

Before task execution, an RGB-D camera detects the mouth position and
two AprilTags. The AprilTag attached to the tray is used to locate the
bowl, while the AprilTag placed at the robot base is used to express the
detected positions in the robot base frame. The mouth position is
measured once before the robot begins moving and remains fixed throughout
the execution.

The motion-planning module uses the learned DMP weights, task-relevant
configurations, detected tray pose, and detected mouth position to generate
the sequential scooping and feeding motions. The generated desired
end-effector configurations are then sent to the low-level controller
for execution by the robotic manipulator and attached spoon.

%%%%%%%%%%%%%%%%%%%%%%%%%%%%%%%%%%%%%%%%%%%%%%%%%%%%%%%%%%%%%%%%%%%%%%%%%%%%%%%%%%%%%%
\subsection{Demonstration and Task Segmentation}
\label{subsec:demonstration_segmentation}

During the demonstration, a human operator performs the scooping and
feeding motions using a spoon instrumented with motion-capture markers.
The motion-capture system records the Cartesian position and Euler-angle
orientation of the spoon, resulting in the demonstrated trajectory
\begin{equation}
    \mathbf{y}^{d}(t)
    =
    \begin{bmatrix}
        \mathbf{p}^{d\top}(t) &
        \boldsymbol{\eta}^{d\top}(t)
    \end{bmatrix}^{\top}
    \in\mathbb{R}^{6},
    \qquad
    t\in[0,T_d],
    \label{eq:complete_demonstration}
\end{equation}
where $T_d$ is the total duration of the demonstration.

The demonstrated trajectory is divided into two motion segments:
scooping and feeding. Let $t_s$ denote the transition instant between
the two segments. The demonstrated scooping and feeding trajectories are
then defined as
\begin{align}
    \mathbf{y}_{\mathrm{scoop}}^{d}(t)
    &=
    \mathbf{y}^{d}(t),
    &&t\in[0,t_s],
    \label{eq:scooping_demonstration}
    \\
    \mathbf{y}_{\mathrm{feed}}^{d}(t)
    &=
    \mathbf{y}^{d}(t),
    &&t\in[t_s,T_d].
    \label{eq:feeding_demonstration}
\end{align}
The initial, transition, and final poses of the demonstrated task are
respectively defined as
\begin{equation}
    \mathbf{y}_{\mathrm{pre}}^{d}
    =
    \mathbf{y}^{d}(0),
    \qquad
    \mathbf{y}_{\mathrm{post}}^{d}
    =
    \mathbf{y}^{d}(t_s),
    \qquad
    \mathbf{y}_{\mathrm{goal}}^{d}
    =
    \mathbf{y}^{d}(T_d),
    \label{eq:demonstrated_task_poses}
\end{equation}
where $\mathbf{y}_{\mathrm{pre}}^{d}$ and
$\mathbf{y}_{\mathrm{post}}^{d}$ denote the pre-scooping and
post-scooping poses, respectively. The post-scooping pose serves as both
the final pose of the scooping segment and the initial pose of the
feeding segment.

The two motion segments are learned independently using the DMP learning
procedure described in Section~\ref{subsec:dmp_learning}, resulting in
the weight matrices
$\mathbf{W}_{\mathrm{scoop}}$ and
$\mathbf{W}_{\mathrm{feed}}$. During task execution, the demonstrated
poses are adjusted according to the detected tray pose and mouth position.
In particular, the positional component of the final feeding pose is
replaced by the initially detected mouth position, while its
Euler-angle orientation is retained from the demonstration.

%%%%%%%%%%%%%%%%%%%%%%%%%%%%%%%%%%%%%%%%%%%%%%%%%%%%%%%%%%%%%%%%%%%%%%%%%%%%%%%%%%%%%%
\subsection{Mouth Perception}
\label{subsec:mouth_perception}

Before task execution, the mouth position is estimated from RGB-D images
captured by a ZED~2i camera. Detailed Expression Capture and Animation
(DECA) is applied to the RGB image to estimate the facial landmarks
\cite{feng2021learning}. The mouth center is determined from the detected
mouth landmarks, while the corresponding depth information is used to
obtain its three-dimensional position in the camera frame, denoted by
${}^{C}\mathbf{p}_{m}\in\mathbb{R}^{3}$.

An AprilTag placed at the robot base is used to determine the
transformation between the camera frame and the robot base frame. The
mouth position expressed in the robot base frame is computed as
\begin{equation}
    \begin{bmatrix}
        {}^{B}\mathbf{p}_{m}\\
        1
    \end{bmatrix}
    =
    {}^{B}\mathbf{T}_{C}
    \begin{bmatrix}
        {}^{C}\mathbf{p}_{m}\\
        1
    \end{bmatrix},
    \label{eq:mouth_position_base}
\end{equation}
where ${}^{B}\mathbf{T}_{C}\in SE(3)$ is the homogeneous transformation
from the camera frame to the robot base frame.

Since only the mouth position is estimated, the final spoon orientation
is retained from the demonstrated feeding motion. Accordingly, the goal
of the feeding DMP is defined as
\begin{equation}
    \mathbf{g}_{\mathrm{feed}}
    =
    \begin{bmatrix}
        {}^{B}\mathbf{p}_{m}\\
        \boldsymbol{\eta}_{\mathrm{feed}}^{d}
    \end{bmatrix}
    \in\mathbb{R}^{6},
    \label{eq:feeding_goal}
\end{equation}
where $\boldsymbol{\eta}_{\mathrm{feed}}^{d}\in\mathbb{R}^{3}$ is the
final spoon orientation extracted from the demonstrated feeding motion.
The mouth position is estimated once before the robot begins moving and
remains fixed throughout task execution.

%%%%%%%%%%%%%%%%%%%%%%%%%%%%%%%%%%%%%%%%%%%%%%%%%%%%%%%%%%%%%%%%%%%%%%%%%%%%%%%%%%%%%%
\subsection{Task Execution Sequence}
\label{subsec:task_execution}

The robotic feeding task is executed in three consecutive phases:
approach, scooping, and feeding. Before the robot begins moving, the tray
AprilTag and mouth position are detected and expressed in the robot base
frame.

The pre- and post-scooping poses are extracted from the demonstration
and represented relative to the tray AprilTag frame $\mathcal{T}$. Let
${}^{\mathcal{T}}\mathbf{p}_{\mathrm{pre}}^{d}$ and
${}^{\mathcal{T}}\mathbf{p}_{\mathrm{post}}^{d}$ denote the demonstrated
pre- and post-scooping positions expressed in the tray AprilTag frame,
respectively. During task execution, the detected tray pose is used to
transform these positions into the robot base frame:
\begin{align}
    \begin{bmatrix}
        {}^{B}\mathbf{p}_{\mathrm{pre}}\\
        1
    \end{bmatrix}
    &=
    {}^{B}\mathbf{T}_{\mathcal{T}}
    \begin{bmatrix}
        {}^{\mathcal{T}}\mathbf{p}_{\mathrm{pre}}^{d}\\
        1
    \end{bmatrix},
    \label{eq:pre_scooping_position}
    \\
    \begin{bmatrix}
        {}^{B}\mathbf{p}_{\mathrm{post}}\\
        1
    \end{bmatrix}
    &=
    {}^{B}\mathbf{T}_{\mathcal{T}}
    \begin{bmatrix}
        {}^{\mathcal{T}}\mathbf{p}_{\mathrm{post}}^{d}\\
        1
    \end{bmatrix},
    \label{eq:post_scooping_position}
\end{align}
where ${}^{B}\mathbf{T}_{\mathcal{T}}\in SE(3)$ is the homogeneous
transformation from the tray AprilTag frame to the robot base frame.

The pre- and post-scooping configurations used during execution are
defined as
\begin{equation}
    \mathbf{y}_{\mathrm{pre}}
    =
    \begin{bmatrix}
        {}^{B}\mathbf{p}_{\mathrm{pre}}\\
        \boldsymbol{\eta}_{\mathrm{pre}}^{d}
    \end{bmatrix},
    \qquad
    \mathbf{y}_{\mathrm{post}}
    =
    \begin{bmatrix}
        {}^{B}\mathbf{p}_{\mathrm{post}}\\
        \boldsymbol{\eta}_{\mathrm{post}}^{d}
    \end{bmatrix},
    \label{eq:pre_post_scooping_configurations}
\end{equation}
where $\boldsymbol{\eta}_{\mathrm{pre}}^{d}$ and
$\boldsymbol{\eta}_{\mathrm{post}}^{d}$ are the demonstrated spoon
orientations at the beginning and end of the scooping segment,
respectively.

In the approach phase, the robot moves from a predefined home
configuration $\mathbf{y}_{h}$ to the pre-scooping configuration
$\mathbf{y}_{\mathrm{pre}}$. The scooping DMP is subsequently initialized
as
\begin{equation}
    \mathbf{y}_{0}
    =
    \mathbf{y}_{\mathrm{pre}},
    \qquad
    \mathbf{g}
    =
    \mathbf{y}_{\mathrm{post}},
    \qquad
    \mathbf{W}
    =
    \mathbf{W}_{\mathrm{scoop}},
    \label{eq:scooping_dmp_parameters}
\end{equation}
where $\mathbf{W}_{\mathrm{scoop}}\in\mathbb{R}^{6\times N}$ is the
DMP weight matrix learned from the demonstrated scooping segment using
\eqref{eq:dmp_weight_learning}, with the weight-matrix structure defined
in \eqref{eq:dmp_weight_matrix}. The scooping DMP then generates the
spoon motion from the pre-scooping to the post-scooping configuration.

After the scooping motion is completed, the feeding DMP is initialized
from the post-scooping configuration toward the feeding goal defined in
\eqref{eq:feeding_goal}:
\begin{equation}
    \mathbf{y}_{0}
    =
    \mathbf{y}_{\mathrm{post}},
    \qquad
    \mathbf{g}
    =
    \mathbf{g}_{\mathrm{feed}},
    \qquad
    \mathbf{W}
    =
    \mathbf{W}_{\mathrm{feed}},
    \label{eq:feeding_dmp_parameters}
\end{equation}
where $\mathbf{W}_{\mathrm{feed}}\in\mathbb{R}^{6\times N}$ is the
DMP weight matrix learned from the demonstrated feeding segment using
\eqref{eq:dmp_weight_learning}.
Using the learned weight matrices and the initial and goal
configurations defined above, the scooping and feeding trajectories are
generated by integrating the DMP transformation system in
\eqref{eq:dmp_transformation_z}--\eqref{eq:dmp_transformation_y},
together with the canonical system in \eqref{eq:dmp_canonical} and the
forcing term in \eqref{eq:dmp_forcing}. The approach trajectory moves
the robot from the home configuration to the pre-scooping configuration.
The complete desired end-effector trajectory is then expressed as
\begin{equation}
    \mathbf{y}_{\mathrm{des}}(t)
    =
    \begin{cases}
        \mathbf{y}_{\mathrm{approach}}(t),
        & t\in\mathcal{I}_{1},
        \\[1mm]
        \mathbf{y}_{\mathrm{scoop}}(t),
        & t\in\mathcal{I}_{2},
        \\[1mm]
        \mathbf{y}_{\mathrm{feed}}(t),
        & t\in\mathcal{I}_{3},
    \end{cases}
    \label{eq:task_execution_sequence}
\end{equation}
where $\mathcal{I}_{1}$, $\mathcal{I}_{2}$, and $\mathcal{I}_{3}$ denote
the approach, scooping, and feeding intervals, respectively. 

%%%%%%%%%%%%%%%%%%%%%%%%%%%%%%%%%%%%%%%%%%%%%%%%%%%%%%%%%%%%%%%%%%%%%%%%%%%%%%%%%%%%%%
\subsection{Robot Control}
\label{subsec:robot_control}

The desired end-effector configuration $\mathbf{y}_{\mathrm{des}}(t)$ generated is then transmitted through ROS~2 as the reference trajectory for a Cartesian impedance controller. The controller tracks the desired position and orientation while providing compliant end-effector behavior during task execution, following the commonly adopted Cartesian compliance framework for physical-interaction tasks \cite{thamrongaphichartkul2024enhancing}. The detected tray and mouth positions are fixed at the beginning of the task and remain unchanged throughout execution.

%%%%%%%%%%%%%%%%%%%%%%%%%%%%%%%%%%%%%%%%%%%%%%%%%%%%%%%%%%%%%%%%%%%%%%%%%%%%%%%%%%%%%%
\section{Experiment}
\label{sec:experiment}

\begin{figure}
    \centering
    \includegraphics[width=\linewidth]{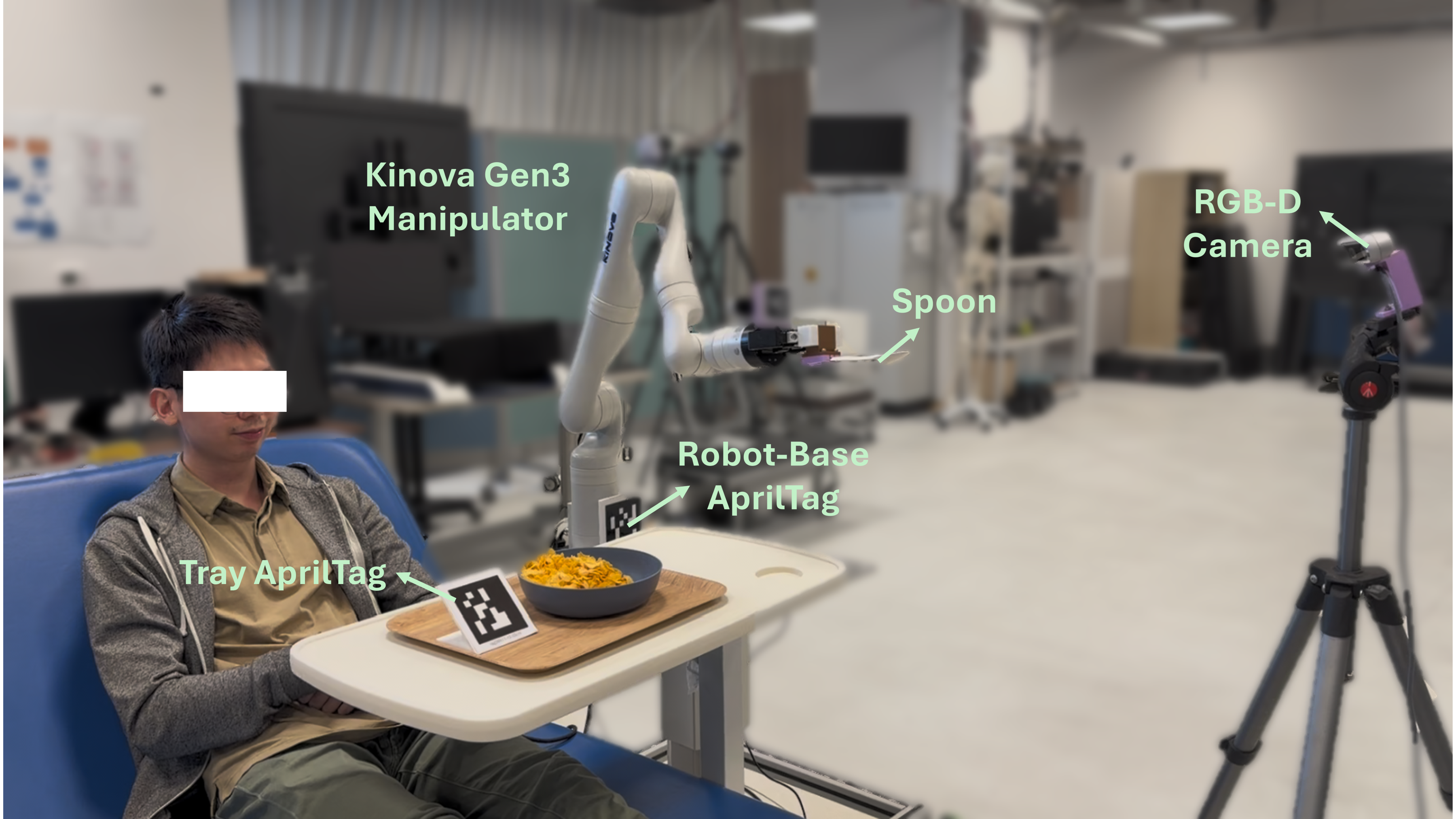}
    \caption{Experimental setup for the robotic feeding task. A Kinova
    Gen3 manipulator equipped with a spoon performs the scooping and
    feeding motions, while an RGB-D camera is used to estimate the
    participant's mouth position. Two AprilTags are used for
    camera-to-robot frame registration and tray localization.}
    \label{fig:experimental_setup}
\end{figure}

\subsection{Experimental Setup}
\label{subsec:experimental_setup}

The experimental setup is shown in
Fig.~\ref{fig:experimental_setup}. The robotic feeding system consists of
a 7-DoF Kinova Gen3 manipulator equipped with a spoon at its end
effector, a ZED~2i RGB-D camera, and a tray containing a food bowl. The
participant is seated in front of the robot, with the tray positioned
within the robot workspace. A reference AprilTag attached to the robot
base is used to determine the transformation between the camera and robot
base frames, while a second AprilTag attached to the tray is used to
localize the tray relative to the robot.

The demonstration used for DMP learning was collected offline using a
PhaseSpace motion-capture system. A single human operator performed one
complete scooping-and-feeding demonstration using a spoon instrumented
with motion-capture markers. The Cartesian position and Euler-angle
orientation of the spoon were recorded to obtain the demonstrated
trajectory $\mathbf{y}^{d}(t)$ described in
Section~\ref{subsec:demonstration_segmentation}. The trajectory was
subsequently divided into the scooping and feeding segments for DMP
learning. The demonstrated pre- and post-scooping positions were
represented relative to the tray frame $\mathcal{T}$ and used for
retargeting during robot execution.

The perception, motion-planning, and robot-control modules are
implemented and integrated using ROS~2. Before each trial, the mouth
position and tray pose are detected once. The resulting mouth position and tray
pose are then used to generate the approach, scooping, and feeding
motions described in Section~\ref{subsec:task_execution} and executed using the Cartesian impedance controller.

%%%%%%%%%%%%%%%%%%%%%%%%%%%%%%%%%%%%%%%%%%%%%%%%%%%%%%%%%%%%%%%%%%%%%%%%%%%%%%%%%%%%%%
\subsection{Experimental Results}
\label{subsec:experimental_results}

\begin{figure}
    \centering
    \subfloat[]{
        \includegraphics[width=0.47\linewidth]{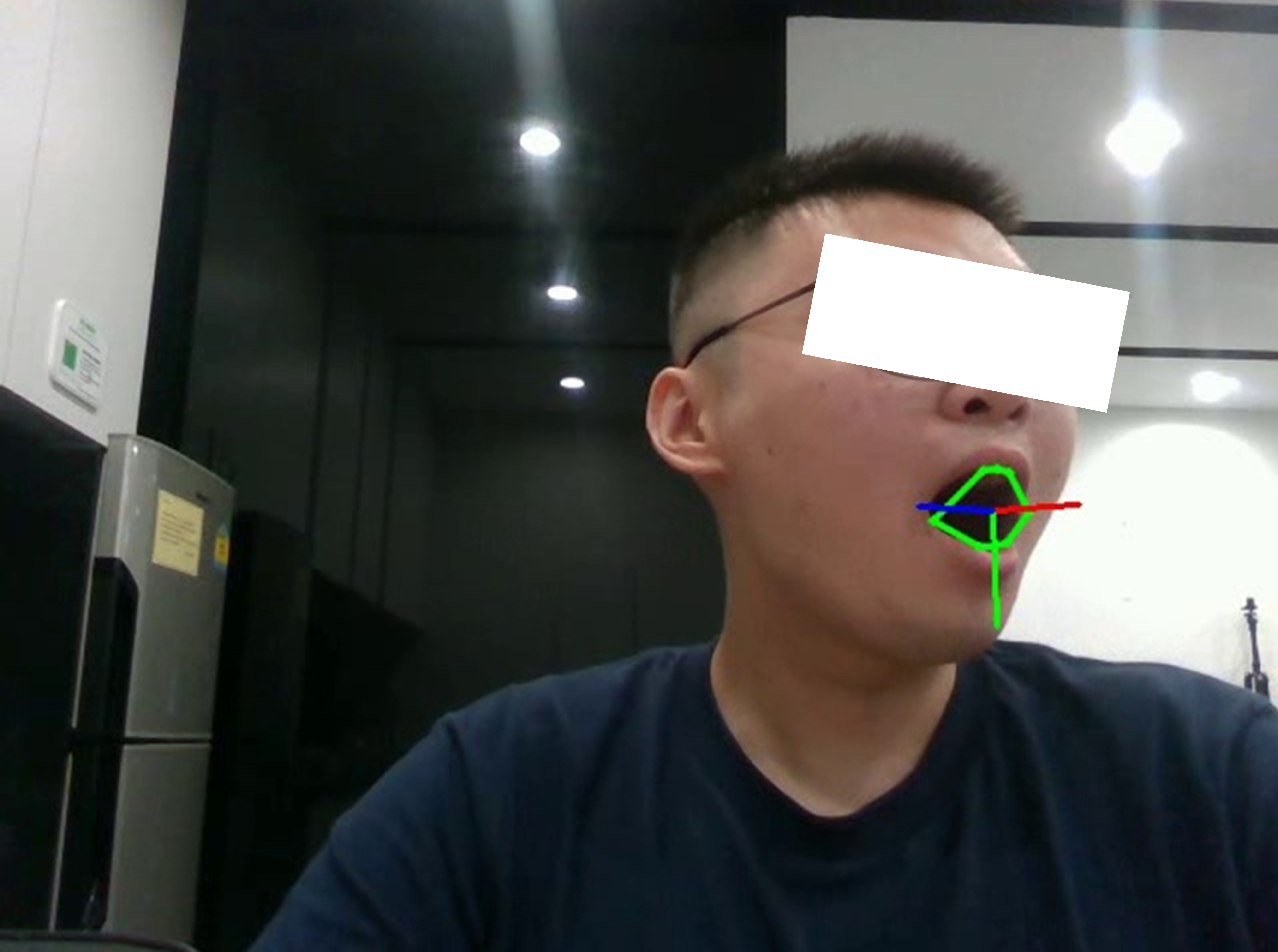}
    }
    \subfloat[]{
        \includegraphics[width=0.47\linewidth]{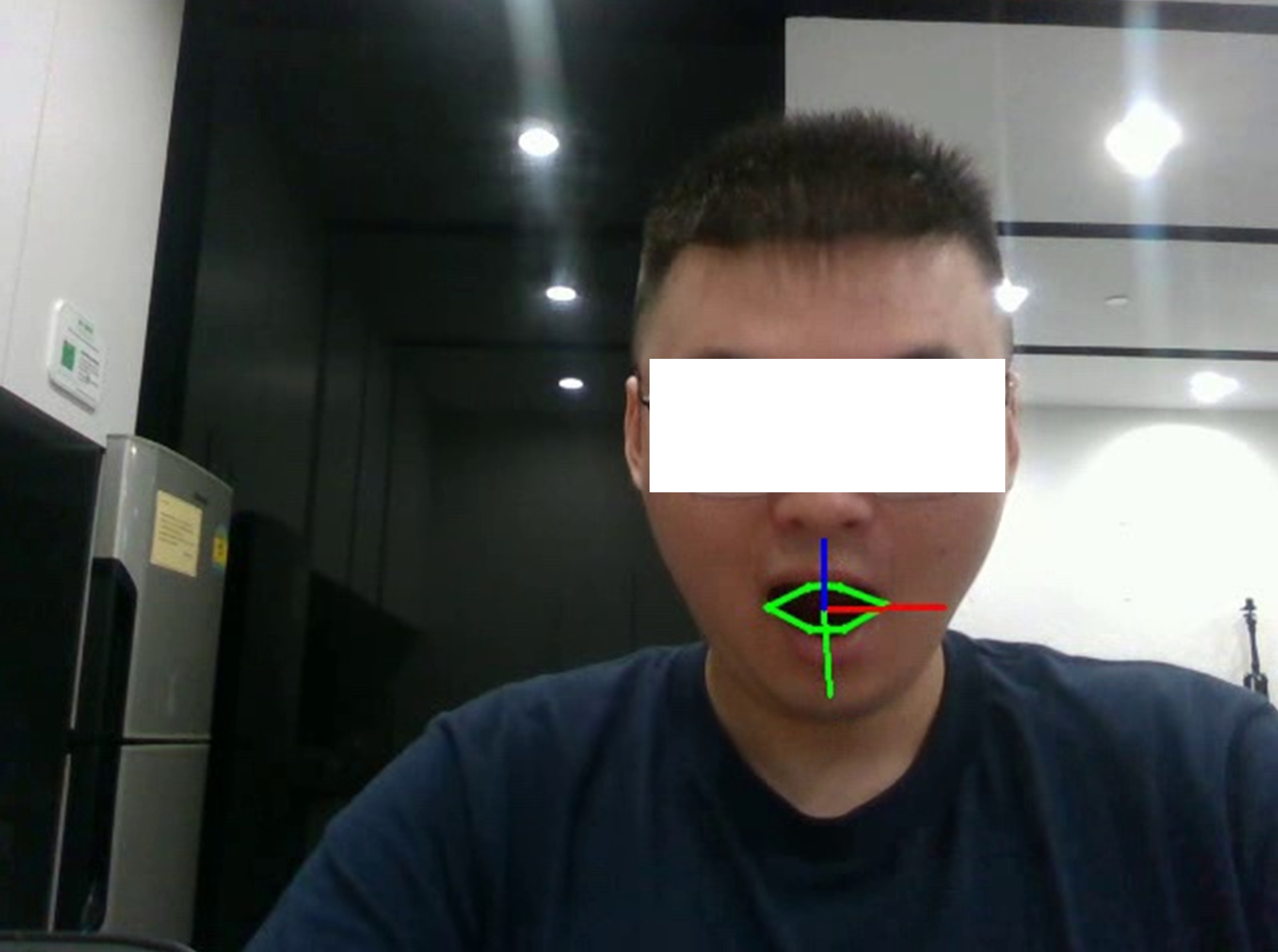}
    }

    \subfloat[]{
        \includegraphics[width=0.47\linewidth]{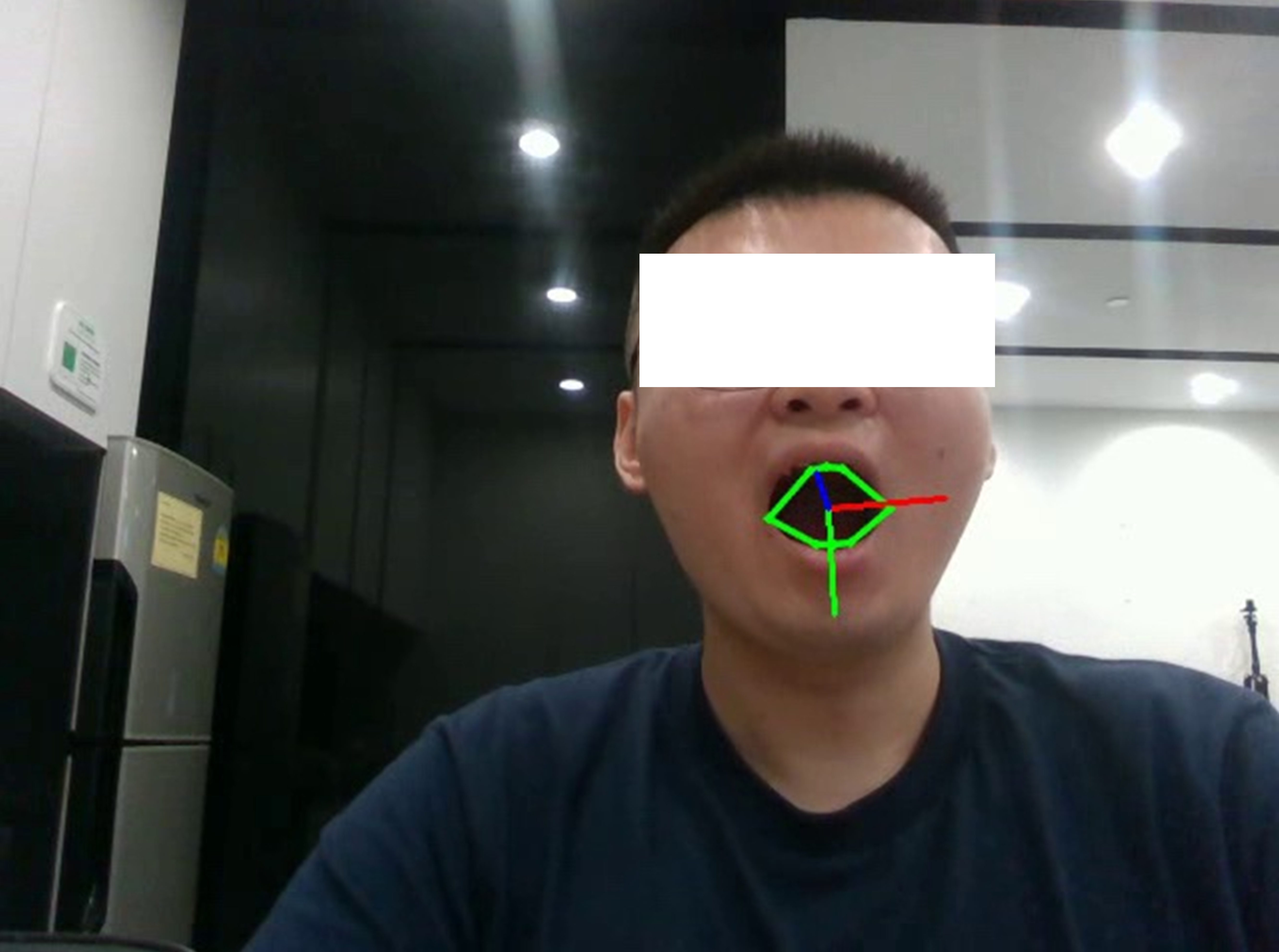}
    }
    \subfloat[]{
        \includegraphics[width=0.47\linewidth]{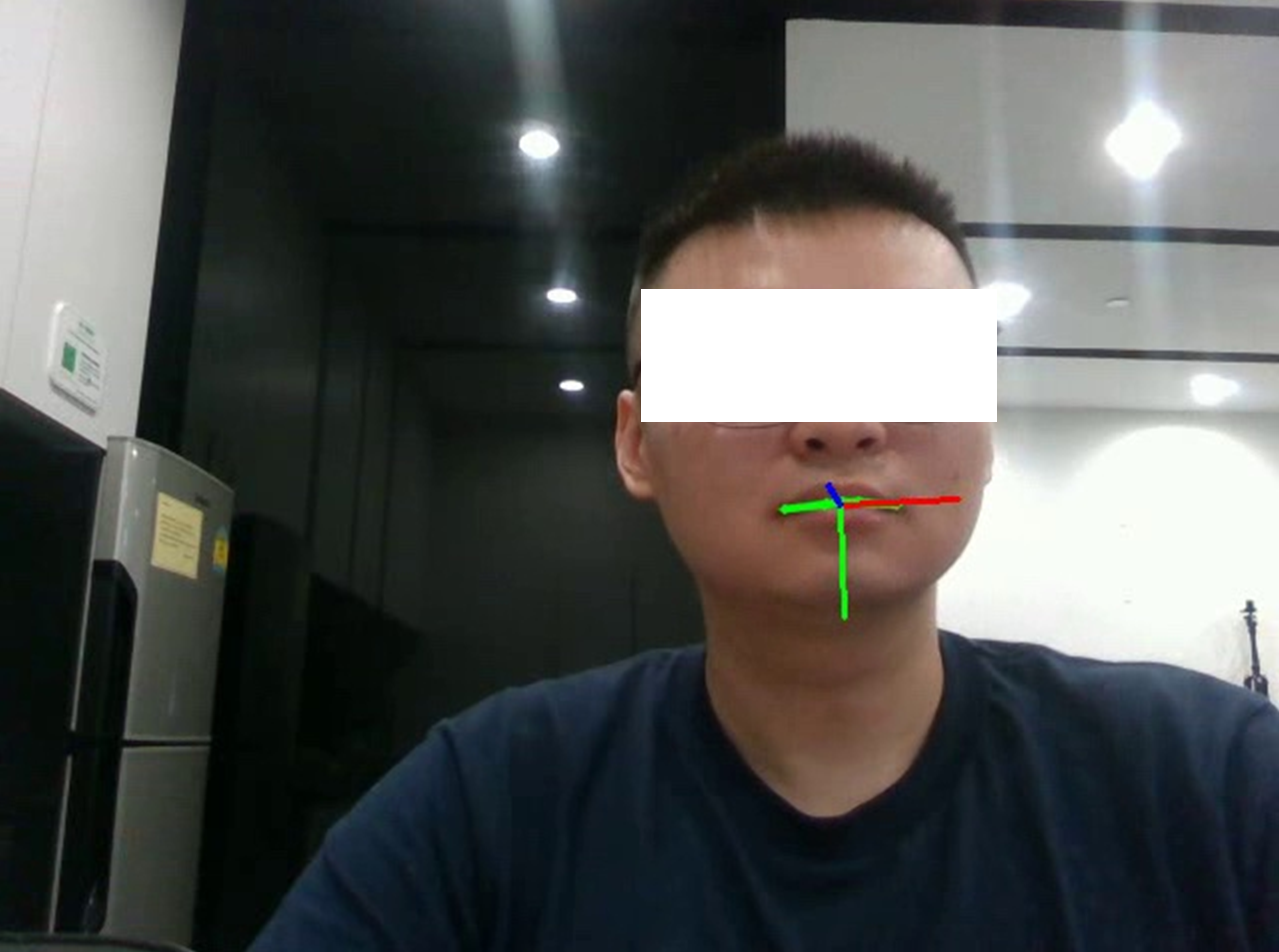}
    }

    \caption{Mouth-perception results under different head
    poses and mouth configurations.}
    \label{fig:mouth_detection}
\end{figure}

\begin{figure*}
    \centering
    \includegraphics[width=\textwidth]
    {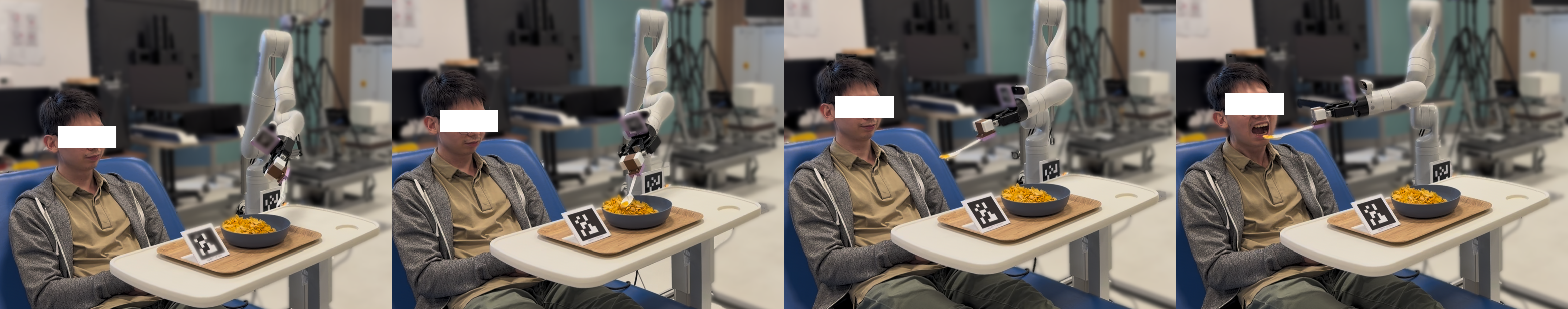}
    \caption{Sequential snapshots of the robotic feeding task. From left
    to right, the manipulator approaches the bowl, executes the learned
    scooping motion, transports the spoon toward the participant, and
    reaches the initially detected mouth position.}
    \label{fig:feeding_execution}
\end{figure*}

\begin{figure}
    \centering
    \includegraphics[width=0.5\textwidth]
    {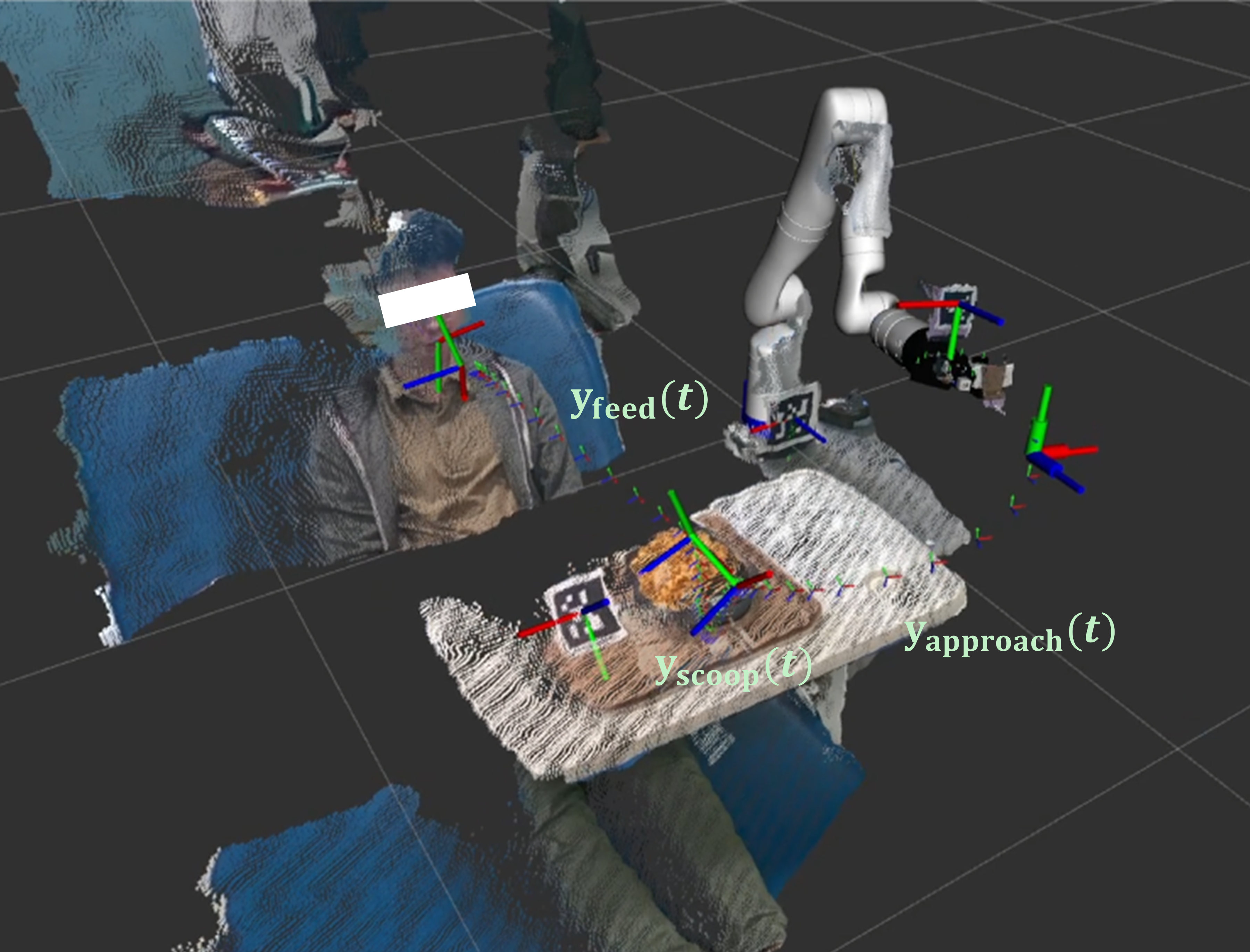}
    \caption{Planned end-effector trajectories visualized within the
    RGB-D point cloud of the experimental scene. The approach trajectory
    $\mathbf{y}_{\mathrm{approach}}(t)$ moves the robot toward the
    pre-scooping pose, while
    $\mathbf{y}_{\mathrm{scoop}}(t)$ and
    $\mathbf{y}_{\mathrm{feed}}(t)$ denote the DMP-generated scooping
    and feeding trajectories, respectively. The coordinate frames
    indicate the planned spoon position and orientation along each
    trajectory.}
    \label{fig:planned_trajectories}
\end{figure}

The experiment began with the demonstration stage, in which a human
operator performed the scooping and feeding motions while the spoon
trajectory was recorded using the motion-capture system. The recorded
trajectory was segmented into the scooping and feeding motions, and the
corresponding DMP weights and task-relevant poses were extracted.

The RGB-D camera was used to estimate the participant's mouth position.
Fig.~\ref{fig:mouth_detection} shows qualitative mouth-perception
results under different head poses and mouth configurations. The
detected mouth landmarks were used to determine the mouth center in the
image, while the corresponding depth information was used to obtain its
three-dimensional position.

The two AprilTags were similarly detected and used for robot-frame registration
and tray localization. The detected tray pose was used to position the
pre- and post-scooping poses, while the detected mouth position was used
as the positional component of the feeding goal. The approach,
scooping, and feeding trajectories were then generated according to the
task sequence described in Section~\ref{subsec:task_execution}.

Fig.~\ref{fig:feeding_execution} shows sequential snapshots of the robot
execution. The manipulator first approaches the bowl, performs the
scooping motion, transports the spoon toward the participant, and
finally reaches the initially detected mouth position.

Fig.~\ref{fig:planned_trajectories} shows the corresponding planned
end-effector trajectories overlaid on the RGB-D point cloud of the
experimental scene. The approach trajectory moves the robot from the
home pose to the pre-scooping pose, followed by the DMP-generated
scooping and feeding trajectories. The coordinate frames plotted along
the trajectories show the planned spoon position and orientation
throughout the task.

%%%%%%%%%%%%%%%%%%%%%%%%%%%%%%%%%%%%%%%%%%%%%%%%%%%%%%%%%%%%%%%%%%%%%%%%%%%%%%%%%%%%%%
\section{Discussion}
\label{sec:discussion}

A key advantage of the framework is that the generated spoon motion
preserves the overall shape and orientation evolution of the
demonstrated motion while allowing its initial and final configurations
to be adapted to the detected tray pose and mouth position. Compared with
direct replay of the recorded trajectory, the DMP representation enables
the demonstrated scooping and feeding motions to be retargeted to the
current experimental scene.

The performance of the framework depends on the accuracy of the RGB-D
mouth localization and AprilTag-based frame registration. Furthermore,
because the mouth and tray poses are estimated only before execution, the
system cannot respond to subsequent changes, such as participant motion
or mouth closure during the feeding motion. Future work will therefore
incorporate continuous mouth perception and online motion adaptation,
including spoon retraction when the mouth closes, to improve the
responsiveness and safety of the feeding task.

%%%%%%%%%%%%%%%%%%%%%%%%%%%%%%%%%%%%%%%%%%%%%%%%%%%%%%%%%%%%%%%%%%%%%%%%%%%%%%%%%%%%%%
\section{Conclusion}
\label{sec:conclusion}

This paper presented a demonstration-based robotic feeding framework
combining human demonstration, DMP-based motion generation, RGB-D
perception, and robot control. Human-demonstrated scooping and feeding
motions were encoded independently and retargeted using the detected
tray and mouth locations. A real-world experiment with a Kinova Gen3
demonstrated the feasibility of executing the complete approach,
scooping, and feeding sequence within the perceived experimental scene.

%%%%%%%%%%%%%%%%%%%%%%%%%%%%%%%%%%%%%%%%%%%%%%%%%%%%%%%%%%%%%%%%%%%%%%%%%%%%%%%%%%%%%%
% \section*{Acknowledgment}

% The preferred spelling of the word ``acknowledgment'' in America is without 
% an ``e'' after the ``g''. Avoid the stilted expression ``one of us (R. B. 
% G.) thanks $\ldots$''. Instead, try ``R. B. G. thanks$\ldots$''. Put sponsor 
% acknowledgments in the unnumbered footnote on the first page.

%%%%%%%%%%%%%%%%%%%%%%%%%%%%%%%%%%%%%%%%%%%%%%%%%%%%%%%%%%%%%%%%%%%%%%%%%%%%%%%%%%%%%%
\printbibliography

\end{document}